\documentclass{bioont}

\usepackage{natbib} 
\usepackage{listings}
\usepackage{tawny}
\usepackage{xspace}
\usepackage[hyphens]{url}

\usepackage{hyperref}

\newcommand{\protege}{Prot\'eg\'e\xspace}
\newcommand{\function}[1]{{\color{blue}\texttt{#1}}}

\lstMakeShortInline[style=tawnystyle]|

\lstnewenvironment{tawny}[1][]%
{\lstset{style=tawnystyle,basicstyle=\scriptsize,frame=single,#1}}{}

\begin{document}

\title{How, What and Why to test an ontology}

\author[J. D. Warrender and P. Lord]{Jennifer D. Warrender and Phillip
  Lord\footnote{To whom correspondence should be addressed:
    phillip.lord@newcastle.ac.uk}}

\address{School of Computing Science, Newcastle University,
Newcastle-upon-Tyne, UK}

\date{March 2015}

\maketitle
\begin{abstract}
  Ontology development relates to software development in that they
  both involve the production of formal computational knowledge. It is
  possible, therefore, that some of the techniques used in software
  engineering could also be used for ontologies; for example, in
  software engineering \textit{testing} is a well-established process,
  and part of many different methodologies.

  The application of testing to ontologies, therefore, seems
  attractive. The Karyotype Ontology is developed using the novel
  Tawny-OWL library. This provides a fully programmatic environment
  for ontology development, which includes a complete test harness.

  In this paper, we describe \textit{how} we have used this harness to
  build an extensive series of tests as well as used a commodity
  continuous integration system to link testing deeply into our
  development process; this environment, is applicable to any OWL
  ontology whether written using Tawny-OWL or not. Moreover, we
  present a novel analysis of our tests, introducing a new
  classification of \textit{what} our different tests are.  For each
  class of test, we describe \textit{why} we use these tests, also by
  comparison to software tests. We believe that this systematic
  comparison between ontology and software development will help us
  move to a more agile form of ontology development.
\end{abstract}

\section{Introduction}
\label{sec:introduction}

Karyotypes have a long history in biology, being used to assess
chromosome rearrangement in many different organisms. In humans, this
knowledge is used diagnostically for many genetic abnormalities. The
use of cytogenetic analysis is cheap, non-invasive and simple, so
remains useful. The representation of karyotypes though, is not
simple. The specification in humans is a hundred-page book, with no
computational definition~\citep{iscn12}. The representation is a
string with no formal grammar which is difficult to manage
computationally.

To address this problem, we have developed the Karyotype Ontology,
which provides a fully computational representation in the form of an
OWL ontology~\citep{warrender-karyotype}.

Ontology development bears many similarities to software development;
both involve taking complex knowledge and producing a computational
amenable representation of that knowledge. For the Karyotype Ontology,
we have extended this similarity further. It has been developed using
Tawny-OWL~\citep{tawny}, a fully programmatic ontology development
environment.

Tawny-OWL is a library, implemented in Clojure which is an
implementation of the Lisp programming language, running on the Java
Virtual Machine. It uses the OWL API~\citep{owlapi}, which is the same
library underlying \protege 4 and upward. It allows constructions of
ontologies programmatically, so rather than adding classes and
properties individually, a large number of entities can be generated
according to patterns defined in
Clojure~\citep{warrender-pattern}. The Karyotype Ontology defines a
number of these patterns, which are used to generate a large number of
classes -- in one case, a single pattern is used to generate 1248
classes. In essence, Tawny-OWL allows us to recast ontology
development as a form of software development, through the use of
functional abstraction.

Tawny-OWL also allows us to use other parts of the software
engineering process; more specifically testing in order to apply
quality control. Historically, ontology testing has been achieved
through the use of DL queries, SPARQL queries and reasoners to ensure
the internal consistency and satisfiability of an ontology. These have
been encapsulated in bespoke tools such as the
efovalidator\footnote{\url{http://www.ebi.ac.uk/fgpt/sw/efovalidator/index.html}}
that can only be used for the validation and unit testing of
EFO~\citep{efo}. More recently, ontology testing has evolved by
incorporating the use of continuous integration systems, as it enables
tests to be run frequently and in a clean
environment~\citep{greycite2899}. Here, the authors support
integration testing while providing releases of OBO ontologies that
are internally consist as well as consistent with external ontologies
and information sources. This
tool\footnote{\url{https://github.com/owlcollab/owltools/tree/master/OWLTools-Oort}}
was initially used to help with the development and maintenance of
GO~\citep{go} and Human Phenotype Ontology~\citep{hpo}, but is not
specific to those domains.

While testing and continuous integration are not novel in the
ontological community, Tawny-OWL has the advantage of not requiring
any specialist installation. Clojure comes with a full test harness,
a build tool for running the tests and is supported by various
continuous integration testing services. In addition, we can use
Tawny-OWL to interact with external ontologies such as GO and OBI. As
tests are simple to use within Tawny-OWL, this has meant that we have
produced a very large test library for the Karyotype Ontology
(currently containing 3088 tests).

In this paper, we describe how we have developed this test suite,
including our use of a spreadsheet to define tests rapidly. We analyse
the different kinds of test and present a novel test classification,
describing the purpose of each form of test. Taken together, this work
represents a systematic attempt to re-purpose software engineering
testing for use within ontology development.

\section{The Karyotype Ontology}
\label{sec:karyotype-ontology}

A karyotype describes the number of chromosomes and any alterations
from the normal. These are visible under the light microscope, and
when stained have a characteristic banding pattern which can be used
to distinguish between different chromosomes and the positions on
these chromosomes.

\begin{sloppypar} 
  Alterations are described by their type, such as inversions,
  deletions or duplications and by their location, specified by a
  chromosome number and band number. So,
  |46,XY,t(1;3)(p22;q13.1)| describes a male with a
  translocation from chromosome |1p22| to chromosome |3q13.1|. To
  describe a karyotype adequately, a unique class needs to be defined
  for each band, of which there are 1224.
\end{sloppypar}

The Karyotype Ontology is developed by specifying the bands in a
literal Clojure data structure, and then using this to generate the
appropriate classes. For example, the following data structure,:

\begin{tawny}
[
 "p10"
 ["p11" "p11.1" "p11.2"]
]
\end{tawny}

describes part of Chromosome 1 which has bands |p10|, and |p11| in
turn has two sub-bands |p11.1| and |p11.2|. The representation was
chosen for ease of legibility/typing. We then use Tawny-OWL to
programmatically expand this data structure into 4 classes, coerced
into a tree, and a set of relationships using code specific to the
Karyotype Ontology.

\section{The Karyotype Test Harness}
\label{sec:test-driv-devel}

Of course, programmers have always tested their code, but
\emph{test-driven} development methodologies emphasise the importance
of writing large numbers of test. A key feature has been the
development of a \emph{tests harness}. This provides a method for
defining tests separate from the main body of code and a mechanism for
running all of the tests regularly in batch. This enables a more agile
form of development, since tests can be run after any change,
detecting if any unexpected changes have occurred.

Clojure provides a test harness which can be used directly with the Karyotype
Ontology. For example, the following statements define two
tests\footnote{Actually, one test with two assertions; the distinction is not
  important in this paper.} which will succeed
if the ontology |human| is both consistent and coherent according to the
reasoner. That is we are asserting that |(r/consistent? human)| returns a
value which |is| true\footnote{The \texttt{r/} part of the statement is the use of a namespace, or namespace alias}.

\begin{tawny}
(deftest Basic
 (is (r/consistent? human))
 (is (r/coherent? human)))
\end{tawny}

These tests can be run either individually or in batch using a single
command. In total, Tawny-OWL itself contains over 3000 individual
assertions. Next, we discuss the kinds of tests that we are running.

\section{The ontology of ontology tests}
\label{sec:test-strat-tawny}

In this section, we use the following terminology to distinguish
between:

\begin{itemize}
\item
  \textbf{tawny-karyotype}\footnote{\url{https://github.com/jaydchan/tawny-karyotype}}\textbf{:}
  the programmatic code written in Clojure, which uses the Tawny-OWL
  library.
\item \textbf{the Karyotype Ontology:} the ontology in OWL, either as
  a set of in-memory Java objects, or as a serialisation as an OWL-XML
  file, which is generated by tawny-karyotype.
\end{itemize}

The first kind of test in tawny-karyotype we describe as
\textbf{software-bound} tests and consists of traditional unit
tests. These are tests where neither the test nor the code that it
tests makes a direct reference to any ontology object. For
example, during the construction of the Karyotype Ontology, it is
useful to be able to determine whether a string, used as a label for a
band, is either on the long (|p|) or short (|q|) arm of a
chromosome. For this purpose, we have defined a \textit{predicate}
function as follows:

\begin{tawny}
(defn str-pband? [band]
 (re-find #"p" band)
\end{tawny}

Here |defn| introduces a function with name |str-pband?| and formal parameter
|band|. This returns true if we |re-find| the regular expression |#"p"| in
|band|. This function is tested against a number of different band labels. The
following examples test that the function returns both true and false
correctly.


\begin{tawny}
(is (h/str-pband? "HumanChromosome1Bandp10"))
(is (not (h/str-pband? "HumanChromosome1Bandq10")))
\end{tawny}

There are 53 of this kind of test. In this case, representative
examples have been generated by hand, and the tests have been directly
written in Clojure.

The second kind of test we call an \textbf{ontology-bound} test, as it
refers to one or more ontology classes or properties. Most of these
use predicates provided by Tawny-OWL or tawny-karyotype. For this
reason, ontology-bound tests are also software-bound. For example, the
following predicate function is defined as part of tawny-karyotype;
this function depends on the |superclass?| function (defined in
Tawny-OWL) and checks to see if |x| is a subclass of
|HumanChromosomeBand|. In this example, |HumanChromosomeBand| is a
term of the Karyotype Ontology, as would be the value passed into |x|.

\begin{tawny}
(defn band? [x]
 (or
  (= x HumanChromosomeBand)
  (superclass? human x HumanChromosomeBand)))
\end{tawny}

This predicate function can then be used to test that the Karyotype
Ontology correctly asserts that the class representing |1p10| is in
fact a chromosome band.

\begin{tawny}
(is (h/band? h/HumanChromosome1Bandp10))
\end{tawny}

There are 759 of this kind of test. As with software-bound tests,
these tests have been written by hand.

The third kind of test, we call a \textbf{reasoner-bound} test as it
uses computational reasoning to determine whether the test passes or
not. All reasoner-bound tests are also ontology-bound. These tests
determine whether the asserted conditions are fulfilled or
not\footnote{Strictly speaking, the \function{band?} function is
  performing a limited, structural form of reasoning by checking
  superclasses recursively.}. As an example, we might make this
assertion, which says that |46,XY| should be a male karyotype.

\begin{tawny}
(is (r/isuperclass? i/k46_XY n/MaleKaryotype))
\end{tawny}

There are 2273 of this kind of test. The majority of these tests are
not directly asserted in Clojure source; we describe how these are
generated in Section~\ref{sec:spec-reas-bound}.

Finally, there is one final type of test which we call
\textbf{probe-bound}. This form of test first changes the ontology in
some way, tests assertions using this changed ontology, and lastly
reverts these changes. Probe-bound tests are also reasoner-bound. In
the following example, we assert a subclass of both |HumanAutosome|
and |HumanSexChromosome|, then define a test assertion that states the
ontology should now be incoherent.  Tawny-OWL provides specific
support for this form of test (|with-probe-entities|), as it is
critical that any entities created during the tests are removed again
to ensure independence.

\begin{tawny}
(is
 (not
  (with-probe-entities
   [_ (owl-class "_"
       :super HumanAutosome
              HumanSexChromosome)]
    (r/coherent?))
\end{tawny}
    
We describe this form of test for completeness, as there are currently
only three of these tests in the Karyotype Ontology. 

\section{Specifying reasoner-bound tests with facets}
\label{sec:spec-reas-bound}

While Tawny-OWL and Clojure provide a reasonably convenient syntax for
specifying most of our tests, it is not ideal for all of them. A large
number of tests for the Karyotype Ontology test the behaviour of a set
of classes which are, effectively, competency questions for our
ontology~\citep{yuan_2014}. The International System for human
Cytogenetic Nomenclature (ISCN) contains a large number of examples
often describing well known conditions or syndromes. These have been
encoded as an ontology as part of tawny-karyotype. The informal nature
of the ISCN as a specification means that these examples are the best
mechanism to ensure that the Karyotype Ontology fulfils the ISCN
specification.


\begin{tawny}
(defclass k45_X
 :super ISCNExampleKaryotype
  (owl-some b/derivedFrom
   (owl-and
    (owl-some b/derivedFrom b/k46_XN)
     (e/deletion 1 h/HumanSexChromosome))))
\end{tawny}

Having defined these example classes it is, of course, useful to test
that they perform as expected when reasoning. We have achieved this by
defining a set of defined classes, which should result in a complex
polyhierarchy after reasoning. We use these classes as facets in a
spreadsheet. Currently, we define 18 facets, with a true/false/unknown
value. For example, |45,X| is defined as \textbf{NOT} male, female or
haploid, but \textbf{IS} diploid, as shown in
Table~\ref{tab:spreadsheet}.

\begin{table}[h!]
  \centering
  \caption{Table showing an excerpt of the ISCN examples facet
    spreadsheet. For each facet we define the value as either: true
    (1), false (-1), or unknown (0).}
  \label{tab:spreadsheet}
  \begin{tabular}{lcccc}
    \hline
    Karyotype & Female & Male & Haploid & Diploid\\
    \hline
    {|45,X|} & -1 & -1 & -1 & 1 \\
    {|45,XX,-22|} & 1 & -1 & -1 & 1 \\
    {|45,X,-X|} & 1 & -1 & -1 & 1 \\
    {|45,X,-Y|} & -1 & 1 & -1 & 1 \\
  \end{tabular}
\end{table}

This spreadsheet is read at test time\footnote{Actually, it is
  translated to a Clojure and is automatically updated when necessary,
  which is a usability and performance enhancement.}  using the
Incanter
library\footnote{\url{https://github.com/incanter/incanter}}. For
example, two of the facets for |45,X| are interpreted as these
assertions:

\begin{tawny}
(is (not (r/isuperclass? i/k45_X n/MaleKaryotype)))
(is (r/isuperclass? i/k45_X n/DiploidKaryotype))
\end{tawny}

The use of a spreadsheet in this way provides a clean and consistent user
interface for specifying facet values. As Tawny-OWL is fully programmatic, it
is straight-forward to store this spreadsheet as part of the source code of
tawny-karyotype which simplifies future updates, and has allowed
us to specify a large number of tests for the Karyotype Ontology. In addition,
this simplifies continuous integration, which we describe next.

\section{Continuous Integration}
\label{sec:cont-integr}

Continuous integration (CI) is a software development process where
code is tested against its dependencies (and code that depends on it)
regularly; in most cases, developers now test their code after every
commit to their version control system. CI provides two key features
in addition to ``normal'' testing. Firstly, it is responsive to
changes in any dependencies, allowing problems to be detected very
early. Secondly, it is normally performed in a ``clean'' environment,
supporting reproducibility.

As tawny-karyotype is using a standard test environment, it is very
easy to set up CI. In our case, we are using
TravisCI\footnote{\url{https://travis-ci.org/}}. By design the
Karyotype Ontology has no dependencies; the CI in this case, tests
against changes in the software dependencies (Tawny-OWL, the OWL API,
HermiT~\citep{hermit}, and Clojure).

\section{Discussion}
\label{sec:discussion}

In this paper, we have described our approach to testing the Karyotype
Ontology. The four different kinds of tests that we describe all have
different purposes. The first of these, software-bound is strictly not a form
of ontology testing at all, but unit testing for the software involved in
ontology development. It is, however, an essential part of our test suite, as
it helps to isolate errors which occur purely as a result of our ontology
development software. Ontology-bound tests directly test our ontology, and
ensure it describes the world correctly -- in essence, they 
are the ontological equivalent of unit tests. The final two forms of tests are
equivalent to functional tests, ensuring the ontology
reasons as we expect. Our taxonomy and test usage differs from previous work
by \cite{ramos-2009}, as we test only T-Box (class) reasoning while they test
the A-Box. In addition, we introduce tests for parts of the infrastructure
outside of the base ontology.

The use of Tawny-OWL has also allowed us to specify tests as facets in a
spreadsheet. Defining a test assertion by filling a cell, means we can test
the karyotype ontology extensively (see Table~\ref{tab:test-stats}). Although,
in our case, we have built the tests using Tawny-OWL with an ontology 
written in Tawny-OWL, it is important to note that the test environment is
de-coupled from the ontology development. Tawny-OWL can use
ontologies written in OWL (by \protege, for instance) and then test them.

\begin{table}[h!]
  \centering
  \caption{Table showing the number of assertions for each test type.}
  \label{tab:test-stats}
  \begin{tabular}{lcccc}
    \hline
    Test Class & Software & Ontology & Reasoner & Probe \\
    \hline
    Base & 0 & 0 & 2 & 0 \\
    Events & 3 & 600 & 2 & 0 \\
    Features & 0 & 0 & 2 & 0 \\
    Human & 50 & 58 & 2 & 0 \\
    ISCN Examples & 0 & 0 & 2156 & 0 \\
    Karyotype & 0 & 1 & 2 & 0 \\
    Named & 0 & 0 & 83 & 0 \\
    Parse & 0 & 28 & 2 & 0 \\
    Random & 0 & 41 & 15 & 3 \\
    Resolutions & 0 & 32 & 7 & 0  \\
    \hline
    Total & 53 & 759 & 2273 & 3 \\
    \hline
  \end{tabular}
\end{table}

We have also briefly described our use of TravisCI, which performs
\textit{integration testing}. The Karyotype Ontology itself has no
ontology dependencies, but we have generated an example ontology which
is a dependency of the Karyotype Ontology and helps to form a test
suite for it. We believe, that the Karyotype Ontology is rather
unusual in having no ontological dependencies. Integration testing is
likely to bear even more fruit for ontologies with a large or complex
dependency graph.

Continuing the metaphor to software engineering, there are currently
several forms of testing that we do not perform on the Karyotype
Ontology. One common problem with ontology development is
understanding reasoner performance, especially the overall reasoning
time. The use of Tawny-OWL does allow performance testing; for
instance, we have extensively compared several different
axiomatisations for parts of the Karyotype
Ontology\footnote{\url{https://github.com/jaydchan/tawny-karyotype-scaling}}.
This form of testing is a \textit{non-functional} test. We do not
currently check overall reasoning performance as part of our automated
test suite, but this is possible and is likely to be included in
tawny-karyotype in the future.

We would also like to test aspects of the ontology other than the
class hierarchy, including extra-logical aspects such as labels or
definitions.  Historically, this form of testing is quite difficult in
Tawny-OWL because we lacked a good mechanism for querying an ontology
syntactically; however, an initial implementation for such a mechanism
(called, prosaically, |tawny.query|) is now in place.

There are a number of tools available for software testing which an equivalent
is not currently available for ontology development within Tawny-OWL, but
which would be extremely useful. We currently, for instance, cannot assess the
state of \textit{coverage} for the Karyotype Ontology as we have neither a tool
nor a clear understanding of how it should assessed for ontologies.

Despite these limitations, the use of Tawny-OWL has allowed us to recast
testing of the Karyotype Ontology as a form of software testing. We have
reused many standard tools to enable this process, and they perform well. In
addition, we have made use of programmatic nature of Tawny-OWL to allow
specification of tests using spreadsheets as source code, using the
extensibility of Tawny-OWL, something we have also found useful for ontology
development. As Tawny-OWL is built on the OWL API, it can offer these
capabilities to any ontology, whatever the development environment. So while
the work reported on here is specific to the Karyotype Ontology, we believe
that the classification of ontology tests and the tooling is generic, and we
look forward the application of these forms of tests to many other ontologies.

\section*{Acknowledgements}

This work was supported by Newcastle University.

\bibliographystyle{natbib}

\bibliography{2015-bio-ont-testable-karyotype}


\end{document}